# TRADEOFFS IN CONSTRUCTING AND EVALUATING TEMPORAL INFLUENCE DIAGRAMS

Gregory M. Provan*
Computer and Information Science Department
University of Pennsylvania
301C 3401 Walnut St.   Philadelphia   PA 19104-6228

## Abstract

This paper addresses the tradeoffs which need to be considered in reasoning using probabilistic network representations, such as Influence Diagrams (IDs). In particular, we examine the tradeoffs entailed in using Temporal Influence Diagrams (TIDs) which adequately capture the temporal evolution of a dynamic system without prohibitive data and computational requirements. Three approaches for TID construction which make different tradeoffs are examined: (1) tailoring the network at each time interval to the data available (rather then just copying the original Bayes Network for all time intervals); (2) modeling the evolution of a parsimonious subset of variables (rather than all variables); and (3) model selection approaches, which seek to minimize some measure of the predictive accuracy of the model without introducing too many parameters, which might cause "overfitting" of the model. Methods of evaluating the accuracy/efficiency of the tradeoffs are proposed.

## 1   INTRODUCTION

This paper examines tradeoffs which need to be considered for reasoning with Probabilistic Networks such as Influence Diagrams (IDs) [16, 26]. For large networks, both data acquisition and network evaluation are expensive processes, and some means of controlling network size is often necessary. In particular, modeling time-varying systems with Temporal Influence Diagrams (TIDs) or Temporal Bayes Networks (TBNs) often requires large networks, especially if several time slices are modeled. We examine three methods of limiting network size, and examine the tradeoffs entailed in each of these methods. Some formal techniques for characterizing such tradeoffs are introduced.

*This work was supported by NSF grant #IRI92-10030, and NLM grant #BLR 3 RO1 LMO5217-02S1.

The main network type examined, the TBN, has been used to model a variety of dynamic processes, including applications for planning and control [11, 12] and medicine (e.g. [2], VPnet [10], and ABDO [24]). In such applications, the static system structure is modeled using a Bayes Network (BN) or influence diagram (ID), and the temporal evolution of the system is modeled using a time series process, connecting nodes in the BN over different time intervals using "temporal arcs". In other words, if $BN_1, BN_2, ...BN_k$ are a temporal sequence of Bayesian networks (called a temporal BN or TBN), these systems address a method of defining the interconnections among these temporally-indexed BNs. The sequence of Bayesian networks (which evolve according to the stochastic dynamic process) together with a corresponding sequence of management decisions and values derived from the decisions defines the temporal influence diagram.

In almost all of these approaches, a Markov assumption is made, due primarily to the entailed well-known theoretical properties and relative computational feasibility. However, this simple form of temporal dependence is violated by many real-world processes. Higher-order Markov processes can be embedded in the TBN or TID to capture longer-term stochastic processes, but at the expense of adding more temporal arcs, thereby increasing data requirements and computational demands of network evaluation.[1] Similarly, other temporal processes, such as dynamic linear models (DLM) [29], can be embedded into temporal BNs or IDs [9, 10, 18].

Some difficulties which arise in large, complicated domains, (e.g. for domains in which large TIDs are constructed [9, 17, 18, 24]), include:

- Given that exact network evaluation is NP-hard [6], and the approximation task is also NP-hard [8], limiting the size of networks is often the only way to ensure computational feasibility. Hence, during model construction, one needs to trade off

---
[1] Modeling time-series processes other then first-order Markov processes can be computationally infeasible for large systems [23].

a utility-maximizing model for parsimony (and computational feasibility).

- It is difficult to evaluate time-series processes for models which contain many variables. In addition, the data collection/storage requirements for large models can be prohibitive.

- Due to certain conditional dependencies among variables, it may make more sense to model the temporal evolution of only the subset of variables which are in fact evolving, and use these processes to drive the changes in the dependent variables.

This paper addresses the tradeoffs inherent in constructing TIDs which adequately capture the temporal evolution of the system without prohibitive data and computational requirements. Three approaches for TID construction which make different tradeoffs are introduced: (1) knowledge-base construction approaches, which tailor the network at each time interval to the data available (rather then just copying the original Bayes Network for all time intervals) [23]; (2) domain-specific time-series approaches, which model the evolution of a parsimonious subset of variables (rather than all variables); and (3) model selection approaches, which seek to minimize some measure of the predictive accuracy of the model without introducing too many parameters, which might cause "overfitting" of the model. The second and third approaches are the main contribution of this paper: the second approach is a new analysis of TIDs, and the third approach is the first application to probabilistic networks of trading predictive accuracy for model parsimony.

The tradeoffs made by these parsimonious approaches are quantified using various methods, and illustrated using a medical diagnosis example. In addition, some Bayesian approaches to model selection are also examined.

## 2  TEMPORAL BAYESIAN NETWORKS

### 2.1  Static Model Structure

We characterize a BN or TID model $\mathcal{M}$ using the pair $(\mathcal{G}, \theta)$, where $\mathcal{G}$ refers to the graphical structure of the model and $\theta$ refers to the set of parameters associated with $\mathcal{G}$, such as conditional probability distributions assigned to arcs in $\mathcal{G}$.

The qualitative structure $\mathcal{G}(V, A)$ consists of a directed acyclic graph (DAG) of vertices $V$ and arcs $A$, where $A \subseteq V \times V$. Each vertex corresponds to a discrete random variable $\psi$ with finite domain $\Omega_\psi$. Arcs in the BN represent the dependence relationships among the variables. Arcs into chance nodes represent probabilistic dependence and are called conditioning arcs. The absence of an arc from node $i$ to $j$ indicates that the associated variable $\psi_j$ is conditionally independent of variable $\psi_i$ given $\psi_j$'s direct predecessors in the DAG $\mathcal{G}(V, A)$.

For a static model (i.e. a single time slice) the quantitative parameter set $\theta$ consists of the conditional probability distributions necessary to define the joint distribution $P(\psi_1, \psi_2, ..., \psi_n)$. The required distributions are given by $P(\psi)$ for every node $\psi$ with no incoming arcs, and by the $P(\psi_i|\psi_j)$ for the nodes $\psi_i$, $\psi_j$ joined by an arc in the DAG. Note that the structure $\mathcal{G}$ unambiguously defines the parameter set $\theta$ which is necessary to specify the joint distribution $P(\psi_1, \psi_2, ..., \psi_n)$, and the structure $\mathcal{G}$ of a BN is implicit in the parametric description.

### 2.2  Example: Acute Abdominal Pain Model

Provan and Clarke [24, 23] have developed an ID model for the diagnosis and treatment of acute abdominal pain (AAP). A common cause of acute abdominal pain is appendicitis, and in many cases a clear diagnosis of appendicitis is difficult, since other diseases such as Non-Specific Abdominal Pain (*NSAP*) can present similar signs and symptoms (findings).

Figure 1: Influence diagram for diagnosis and treatment of acute abdominal pain

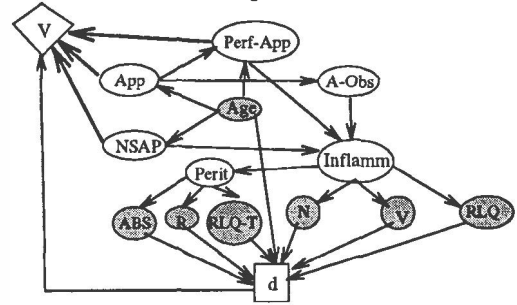

In this model, a BN models the physiology of the system, and decision and value nodes represent the actions taken and corresponding utilities of such actions respectively. Figure 1 presents an example of the type of network created for the diagnosis of AAP for a single time slice. In this figure, chance, decision and value nodes are represented diagrammatically by ovals, rectangles and diamonds respectively. For example, the chance node for Inflammation (*Inflamm*) is conditionally dependent on the chance nodes for Perforated-Appendix (*Perf-App*) and Appendicial-Obstruction (*A-Obs*). Some possible diseases studied in the model are Appendicitis (*App*) and *NSAP*. In this single time-slice ID there is one decision node $d$ and one value node $V$. The shaded nodes in this diagram represent observable variables $\mathcal{X}$, e.g. Absent Bowel Sounds (*ABS*), Right-Lower-Quadrant Tenderness (*RLQ-T*), Nausea (*N*), Vomiting (*V*), etc.

### 2.3  Dynamic Model Structure

A temporal BN (or ID) consists of a sequence of BNs (IDs) indexed by time, e.g. $\mathcal{G}_0, \mathcal{G}_1, ..., \mathcal{G}_t$, such that temporal arcs connect $\mathcal{G}_i$ with $\mathcal{G}_j$, with the direction of the arcs going from $i$ to $j$ if $i < j$.[2] A *temporal arc* $A_\tau(t)$ connecting networks for time slices $t-1$ and $t$ is a subset of the inter-network edge set $A_{int}(t)$, given by

$$A_{int}(t) = \{(\alpha,\beta)|\alpha \in V(t-1), \beta \in V(t),$$
$$(\alpha,\beta) \in V(t-1) \times V(t)\}.$$

If we index the BN by time, i.e. $\mathcal{G}_t = (V(t), A(t))$, then the full temporal network over $N$ time slices (which may be intervals or points), is given by $\mathcal{G}^N = (V^N, A^N)$, where

$$V^N = \bigcup_{t=0}^{N} V(t), \text{ and}$$
$$A^N = \bigcup_{t=0}^{N} A(t) \cup \bigcup_{t=1}^{N} A_\tau(t).$$

Each temporal arc connects a pair of vertices

$$\mathbf{v}^p(t) = \{v_i(t-1), v_j(t)|(v_i(t-1), v_j(t)) \in A_\tau(t)\}.$$

The temporal node set connected over time slices $t-1$ and $t$ is given by

$$\mathbf{V}^p(t) = \{V_i(t-1) \cup V_j(t)|A_\tau(t) \subset V_i(t-1) \times V_j(t)\}.$$

### 2.4 Example: Temporal Model for Diagnosis

Temporal reasoning for AAP is important due to the difficulty of diagnosis and treatment based on data from just a single time slice. Appendicitis progresses over a course of hours to days, and one might be tempted to wait until the complex of signs and symptoms is highly characteristic of appendicitis before removing the appendix. However, the inflamed appendix may perforate during the observation period, causing a more generalized infection and raising the risk of death from about 1 in 200 cases to about 1 in 42 [21]. Thus, the tradeoff is between the possibility of an unnecessary operation on someone whose findings are similar to early appendicitis and a perforation in someone whose appendicitis is allowed to progress.

Given that data over time can greatly simplify the diagnostic process, a TID is used for this domain. As an example, consider a simple situation in which 2 temporal intervals are modeled for the AAP domain as shown in Figure 2. Dashed lines indicate the arcs joining nodes from two different time slices.

### 2.5 Parametric Specification

The probability distributions to be specified for a TID can be classified into two types: (1) time-series process distributions for temporal arcs $A_{int}(t)$; and (2) distributions for the network for each time slice, $\mathcal{G}_1, \mathcal{G}_2, ....$ For example if Figure 2 represents graphs for two time

---

[2]This notation is adapted from [18].

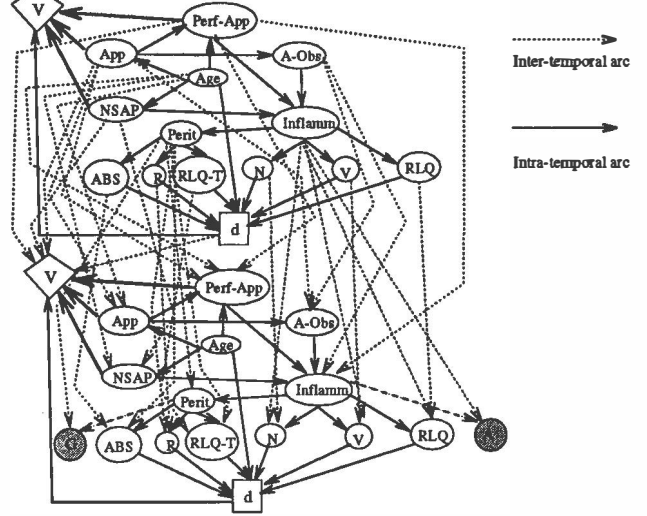

Figure 2: TID for patient $X$ over 2 time intervals, with new findings of anorexia $(A)$ and muscular guarding $(G)$ in the second time interval, as shown by shaded nodes

slices, $\mathcal{G}_1$ and $\mathcal{G}_2$, then a sample of temporal arc distributions includes: $P(V(2)|V(1))$, $P(RLQ(2)|RLQ(1))$, $P(ABS(2)|ABD(1))$, $P(App(2)|App(1))$, $P(Perf - App(2)|Perf - App(1))$, etc. A sample of distributions within a single time slice includes: $P(V(1)|Inflamm(1))$, $P(RLQ(1)|Inflamm(1))$, $P(ABS(1)|Perit(1))$, $P(App(2)|App(1))$, $P(Inflamm(1)|NSAP(1))$.

## 3   TID CONSTRUCTION FROM KNOWLEDGE BASES

TIDs (or TBNs) are typically constructed (e.g. [11]) by replicating the ID for the initial time slice over the succeeding $N-1$ time slices (i.e. $\mathcal{G}_i, i = 0, 1, ..., N$ are all the same), and then joining the networks over successive time slices using a Markov assumption. This approach is relatively inflexible, as it does not allow the network to be altered over time. In addition, if the network's size changes over time, many redundant variables will be present as the static network is replicated for future time slices, since the first network will need to incorporate all potentially relevant structure over future time slices.

A recent approach to reduce (static) Bayes network complexity, tailoring networks to data [15, 28], offers the potential to improve network evaluation costs for such networks. This approach entails constructing a knowledge base (KB) for the domain in question. Given a particular set $O$ of observations, this approach does not construct a network corresponding to the entire KB, but instead tailors a model to the observations $O$ from the KB.

This approach has been extended to the construction of TIDs in [24], where a first-order Markov assumption was made in defining the temporal arcs $A_\tau(t)$. For example, Figure 2 represents a TID for two time slices. Note that even though the KB for the acute abdominal pain domain covers over 50 findings, 20 intermediate disease states and 4 diseases [24], the network $\mathcal{G}_1$ for the first time slice is significantly simpler, and covers only 7 findings, 4 intermediate disease states and 2 diseases.[3] Evaluating this smaller network can be done much more efficiently. Note also that this approach can model how the findings change over time during the evolution of the underlying disease by altering the $\mathcal{G}_i$'s over time. For example, in Figure 2, the network $\mathcal{G}_2$ for the second time slice introduces variables not contained in $\mathcal{G}_1$, representing findings present in time slice 2 but absent in time slice 1.

For complex domains like the diagnosis of AAP, the reduction in network size afforded by the automatic network construction approach improves computational efficiency, but not enough to allow modeling complex time-series processes like higher-order Markov processes. Some other techniques, such as the one discussed below, are also necessary.

## 4 DOMAIN-SPECIFIC TIME-SERIES MODELS

In this section we propose two new domain-specific heuristics for cases in which even tailoring the network to the observations does not produce an easily-evaluated BN or ID. For TIDs, a promising heuristic is to model the temporal evolution of only a subset of variables. Two different models for which variables should evolve are possible: "driving" variables or observable variables. These are discussed below.

### 4.1 Parsimonious Modeling of System Temporal Evolution

**Driving Variables:** This approach entails a domain-dependent identification of the system variables which are actually evolving, driving changes in other system variables. To this effect, we partition the system variables $\psi$ into a set $\mathcal{D}$ of dynamic or evolving variables and a set $\mathcal{S}$ of variables which are either constant or whose changes are due to some $x \in \mathcal{D}$. For completeness, we assume a set $\gamma$ of variables which are independent of the variables $x \in \mathcal{D}$. Under this partition, we have $\psi = \mathcal{D} \cup \mathcal{S} \cup \gamma$.

Using this partition, an appropriate stochastic process is associated with each $x \in \mathcal{D}$. In a TID, this is represented by an appropriate set of temporal arcs for each such stochastic process.

This partition should be made to trade off model accuracy for computational efficiency. In some domains, there may be techniques to govern which variables may be modeled as static, and which must be dynamic. In other domains, in order to make the appropriate tradeoffs, heuristics must be used. Section 5 presents some ways to formally evaluate the tradeoffs which are made.

**Observed Variables:** This approach seeks to model the observables (findings) $\mathcal{X}$ which are the evidence of the internal evolution of the system. Typically, when one is monitoring the system, there exists data (over time) for these variables. However, if these variables are not the ones that are driving the process under study, then one is estimating the values of the driving variables $\mathcal{D}$ from the observables, using the model to relate the two classes of variables.

We now present an example of the modeling of acute abdominal pain to demonstrate this temporal arc selection process.

### 4.2 Example: Acute Abdominal Pain Model

The AAP model has three variable types: observable, intermediate (latent) and disease variables, denoted $\mathcal{X}, \mathcal{V}, \mathcal{W}$ respectively. The current method for modeling AAP over time is to use a TID in which a semi-Markov process governs the evolution of *all* system variables [23].[4] This entails defining a large number of temporal arcs. Figure 2 shows a simple situation in which 2 temporal intervals are modeled. For just a first-order Markov assumption, the large number of temporal arcs in Figure 2 is immediately obvious. Model evaluation is consequently very expensive. More importantly, the true temporal processes for this domain are not adequately captured by this first-order Markov assumption [22]. Hence, a higher-order Markov model is required to capture this more complex system evolution. However, this should be embedded without introducing significantly more temporal arcs (with their entailed data and computational-resource requirements).

One solution is to model the evolution of a subset of the variables. Two approaches to developing a model in which only a subset of the variables evolve over time include:

**Driving Variables:** This approach models the underlying physiology which is driving the evolution of the system. Consider the set of causal relationships:

$$\text{App} \rightarrow \text{A-Obs} \rightarrow \text{Inflamm} \rightarrow \text{V}.$$

If we assume that once a case of appendicitis is initiated by an appendicial obstruction (A-Obs), and that the obstruction does not change, then the only variable which changes is the inflammation. Vomiting (V) changes in response to the degree of inflammation.

---

[3] A network like this is constructed for a particular case in which 7 findings are presented.

[4] In this semi-Markov model, data is available to estimate transition distributions for the findings variables, and transition distributions are estimated based on expert opinion for the remaining variables. Dr. J.R. Clarke is the surgeon providing the expert opinion.

A similar analysis can be done to identify the set of variables $\mathcal{D}$ which drive the system evolution. This partitions the variables into static $\mathcal{S}$ and dynamic $\mathcal{D}$ variables: $\psi(t) = \mathcal{S} \cup \mathcal{D}(t)$. The set $\mathcal{D}(t)$ consists of latent variables, as shown in Figure 3. The findings, which also change over time, are conditionally dependent on the $x \in \mathcal{D}(t)$, and are an observable reflection of the internal physiological changes over time.

Figure 3: Reduced version TID for patient $X$ over 2 time intervals, with just disease and latent variables evolving temporally

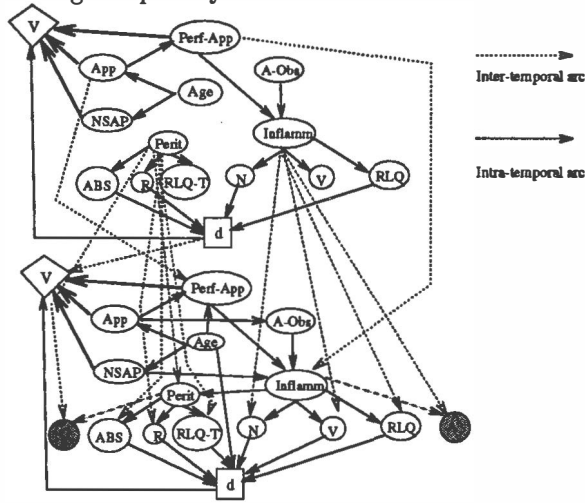

The drawback to this approach is that $\mathcal{D}(t)$ consists of latent variables, for many of which detailed temporal physiological models do not exist. For example, insufficient information about the progressive inflammation of the appendix is known to create a parameterized model, nor can direct measures of the degree of inflammation be made, except possibly using white blood count (WBC); instead, this process is typically inferred from the findings which accompany it.

**Observable variables:** This approach models the observables (findings) $\mathcal{X}$ which are the evidence of the internal evolution of the system. A large body of data exists for these variables [24], as data collection is simplest for these variables. An example of such a network is shown in Figure 4.

The drawback to this approach is that the observables $\mathcal{X}(t)$ may not necessarily predict the underlying diseases $\mathcal{W}(t)$ as reliably as the latent variables $\mathcal{V}(t)$, if the latent variables $\mathcal{V}(t)$ are assumed to be static.[5] A second drawback is that there are relatively more observable than finding variables, so this approach is more computationally expensive than using latent variables alone.

---

[5] A more accurate model might include both latent and finding variables as dynamic variables.

Figure 4: Reduced version TID for patient $X$ over 2 time intervals with temporal arcs for both findings (observable variables) and latent variables

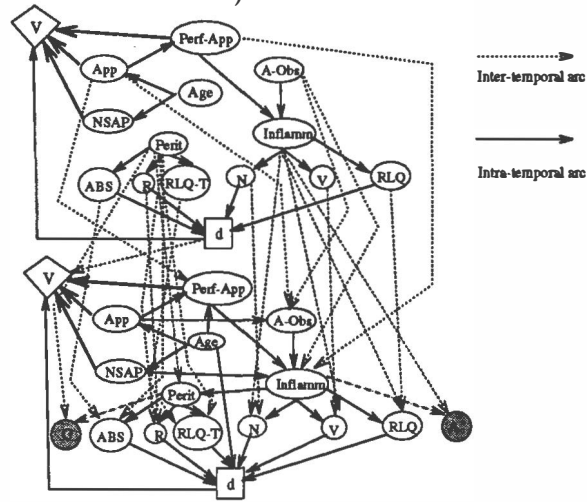

## 5 MODEL SELECTION APPROACHES

### 5.1 Specifying Tradeoffs

If the topology and probabilities of the TID are changed, then some measure of how the changes affect the predictive accuracy of the output, or of the "quality" of the decision-making provided by the network, needs to be computed. In this analysis, we assume that there is some utility function which is used as a measure the network "accuracy" or "decision quality" for different models.[6]

The effectiveness of any decision rule is measured using a loss function $L(\theta, \delta)$. This is interpreted as measuring the loss $L(\theta, \delta)$ associated with taking action $\delta$ when the world state is parameterized by $\theta$. Given a prior probability estimate $\pi(\theta)$ of the world state $\theta$, the *risk function* provides a measure of the expected loss under varying values of the observable variables $\mathcal{X}$, and is given by $R(\theta, \delta) = E_\theta^{\mathcal{X}}[L(\theta, \delta(X))]$.

The *Bayes risk* of a decision rule $\delta$, with respect to a prior distribution $\pi$ on the entire parameter space $\Theta$, is defined as $r(\pi, \delta) = E^\pi[R(\theta, \delta)]$. This averages over the risk functions given all priors that can be assigned to $\theta$.

A decision rule $\delta_1$ is preferred to a rule $\delta_2$ if $r(\pi, \delta_1) < r(\pi, \delta_2)$. A *Bayes rule* is a decision rule which minimizes $r(\pi, \delta)$, and is thus optimal.

This paper proposes a variety of techniques for ana-lyzing the tradeoffs made during model selection, including risk-based as well as purely probabilistic criteria. Although a utility measure is desired, probabilistic criteria can be used in a variety of situations (e.g. [4, 27]). Many probabilistic criteria are simpler to compute, and do not require a prior distribution $\pi(\theta)$.

---

[6] We use a loss function, to maintain consistency with much of the decision theory literature (e.g. [3]); utility and loss, for the purposes of this paper, are duals to each other. Hence, one seeks either to maximize the expected utility, or minimize the expected loss.

For example, in the medical example described earlier, the utility function measures the utility of the treatment given what disease is actually present. Thus an unnecessary appendectomy will have low utility, and a necessary appendectomy will have relatively high utility. So if the loss $L(\theta_1, \delta_1)$ associated with decision $\delta_1$ under parameter set $\theta_1$ is less than that under model $\theta_2$, i.e. $L(\theta_1, \delta_1) < L(\theta_2, \delta_1)$, this means that model $\theta_1$ allows you to provide better treatment under the same decision rule $\delta_1$ than $\theta_2$. For the purposes of medical treatment one needs to determine if the difference is significant.

In addition to the loss function, one may want to trade off decreased model utility for increased computational efficiency. The model parameter penalty $g(\gamma)$ to be introduced in equation 1 can be used to provide a measure of computational expense based on the number of model parameters. Alternatively, one can incorporate into the loss function a computation penalty function $\kappa(\theta)$, which measures the computational resources necessary to evaluate a model with parameters denoted by $\theta$.

### 5.2  Statistical-Estimation Model Selection Approaches

Consider the case where there are $p$ total observable parameters (predictors) $\{\theta_1, \theta_2, ..., \theta_p\}$, of which some subset $q < p$ is to be selected to estimate a latent variable $y$. Possible measures of predictive accuracy based on a parameter estimate $\hat{\theta}$ are:

| | |
|---|---|
| Sum of Squared Error (SSE) | $\|\theta - \hat{\theta}\|^2$ |
| Log likelihood | $log(\theta - \hat{\theta})$ |
| Predictive Risk | $E_\theta\|\theta - \hat{\theta}\|^2$ |

Corresponding to the $p$ parameters, we introduce a set of $p$ indicator variables given by $\gamma = \{\gamma_1, \gamma_2, ..., \gamma_p\}$, where $\gamma_i$ is defined as follows:

$$\gamma_i = \begin{cases} 0 & \text{if } \theta_i \text{ is to be estimated by } 0 \\ 1 & \text{otherwise.} \end{cases}$$

Define $\Gamma$ as the set of all $p$-tuples $\gamma$. $\gamma$ can be thought of as an indicator vector denoting which parameters are considered in a model, and $\Gamma$ as the set of all possible models over $\theta$.

The model-selection procedure then consists of selecting some $\gamma \in \Gamma$ and then estimating $\theta$ by $\hat{\theta}_\gamma$.[7] Various criteria for this process have been used to compute the quality of the model for prediction. The selection procedure criteria can be defined using the following equation:

$$\Upsilon(\theta, \gamma) = f(\theta, \gamma) + g(\gamma), \qquad (1)$$

where $f(\theta, \gamma)$ is a measure of predictive error, and $g(\gamma)$ is a penalty for the number of model parameters. One widely-studied approach is to choose some $\gamma_\Pi \in \Gamma$ that jointly minimizes the sum of predictive error[8] and parametric penalty, setting the predictive error measure to be the sum of squared error (SSE) [13]:

$$\gamma_\Pi = \arg\min_{\gamma \in \Gamma}[SSE_\gamma + |\gamma|\sigma^2\Pi], \qquad (2)$$

where $\Pi \geq 0$ is a pre-specified constant, $|\gamma|$ is the number of nonzero components of $\gamma$, and $SSE = |\hat{\theta}_\gamma - \theta|^2$. In the right-hand-side of equation 2, the first term denotes the predictive error, and the second term is a penalty function on the number of parameters in the model. Hence this equation can capture a wide varieties of approaches which trade off predictive accuracy and model size. For example, for known $\sigma^2$, the Akaike Information Criterion (AIC) [1] is the special case of $\gamma_\Pi$ when $\Pi = 2$, and the BIC approach [25] is the special case of $\gamma_\Pi$ when $\Pi = logn$. A third approach, called the *risk inflation* (RI) approach [13], is defined with respect to a "correct" model parameter set $\gamma^*$ (e.g. as determined by an oracle). The risk inflation measure $RI(\gamma)$, is

$$RI(\gamma) = \sup_\theta \frac{E_\theta|\theta - \hat{\theta}|^2}{E_\theta|\theta - \hat{\theta}_{\gamma^*}|^2} = \frac{R(\theta, \hat{\theta})}{R(\theta, \hat{\theta}_{\gamma^*})}.$$

The selection procedure with smallest risk inflation will be minimax with respect to the ratio function $RI(\gamma)$ [13]. The risk inflation criterion calibrates the risk of a model selection estimator against the risk of an ideal model selection procedure.

### 5.3  Bayesian Model Selection Approaches

The Bayesian approach to model selection is based on computing the posterior probabilities of the alternative models, given the observations. Two Bayesian analyses of the model selection process applied to BNs have been published recently. One method focuses on averaging over all possible BN models to select a model with improved predictive ability [20, 19]. Since the space of all possible models is potentially enormous, two approximation techniques are proposed: (1) use Markov chain Monte Carlo simulation to directly approximate the model selection process [20]; and (2) select a subset of the set of all models by excluding all models which receive less support from the data then their simpler (in terms of number of parameters) counterparts [19]. Both studies indicate that model averaging improves predictive performance.

---

[7]Details for computing $\hat{\theta}_\gamma$ are given in [22].

[8]$\sigma^2$ denotes the variance of the random predictive error in the estimation process.

Given a large model space (as denoted by $\Gamma$), the model space pruning heuristics proposed in [19] can be crucial to the model selection process, given no prior knowledge about alternative models. In contrast, here we present model selection techniques which are useful when a small set of alternative models is being considered (i.e. the entire model space is not considered).

A second approach examines BN structure purely from the viewpoint of predictive accuracy [7]. This approach computes a logarithmic score for alternative models, ignoring the number of parameters in the model. Given a discrete random variable $y$ whose value is to be estimated from a model denoted by the parameter set $\theta$, the scoring rule used is $-logP(y|\theta)$. This approach is thus a restriction of equation 1 to the case where $f(\theta, \gamma) = -logP(y|\theta)$, and $g(\gamma) = 0$. In addition, this selection process is sequential, in that scores are summed over a set of $M$ cases: if $S_m = -logP_m(y|\theta)$ is the score on the $m^{th}$ case, the total score for a particular model is given by

$$S = \sum_{m=1}^{M} -logP_m(y|\theta).$$

This approach allows monitoring of the performance of models as new data becomes available (by updating the score $S$), facilitating model adaptation over time. Several related scoring rules are also analyzed in [7].

## 6    EVALUATING TRADEOFFS

We now present results from a simple AAP pilot study which applies these different model selection criteria to a set of models. In the following, we assume we know the true state of the world, as represented by the canonical model $\theta^*$. The goal is to compare to $\theta^*$ alternative models $\theta_i, i = 1, ..., k$, where the models differ by the time-series process parameters for temporal arcs $A_{int}(t)$.

The BN model analysed is a network consisting of 5 copies of the BN portion of the ID presented in Figure 1, joined together by temporal arcs based on four temporal models: (1) $1^{st}$-order Markov; (2) $2^{nd}$-order Markov; (3) driving parameters; (4) observable parameters.[9] The canonical model was assumed to be the $1^{st}$-order Markov model, even though the long-term nature of the disease evolution may violate this $1^{st}$-order Markov assumption. This choice was made because this model has been studied most carefully to date.

Measures for these models were computed using four different criteria: (1) AIC; (2) BIC; (3) Risk Inflation; (4) BIC with $\Pi = 0$.[10]

Due to space limitations, the full details of this pilot analysis are omitted. A summary of the results is as follows. The AIC criterion selected the $1^{st}$-order Markov model. The BIC and Risk Inflation criteria selected the observable parameter model. The BIC with $\Pi = 0$ criterion selected the $2^{nd}$-order Markov model by a narrow margin over the canonical model; without a penalty for model size this criterion suggests that a $2^{nd}$-order model may actually best fit this data. In contrast, imposing a penalty for model size on this BIC test selects a simpler model, indicating that the cost of adding parameters for the $2^{nd}$-order model outweighs the increased predictive accuracy (given the chosen penalty $\Pi$).

Although this analysis is informative, further analysis is clearly necessary. The selection of the observable parameter model over the driving parameter model may be due to the availability of better data for the observable parameters than the latent parameters.[11] Further, this analysis needs to be done for a large number of cases; however, this pilot study has shown the promise of these model evaluation criteria.

## 7    RELATED LITERATURE

The methods of analyzing networks presented here are orthogonal to the approach proposed by Goldman and Breese [14]. Goldman and Breese describe methods of integrating model construction and evaluation during the process of automated network construction. The main thrust of the work presented here is examining alternative network structures. However, some of the model selection criteria examined here can be used during automated network construction to provide scoring rules for whether nodes and/or arcs should be added to a partially-constructed network.

Of the work in temporal probabilistic networks, the most closely associated work is that of Dagum et al. [10, 9]. The system proposed in [10] is primarily interested in the statistical process underlying temporal Bayesian networks. To this end, the paper focuses on computing inter-temporal conditional dependence relations; in other words, if $BN_1, BN_2, ...BN_k$ are a temporal sequence of Bayesian networks, Dagum et al. address a method of defining the interconnections among these temporally-indexed BNs. In [9] an additive BN approximation model is proposed. Parameter estimation is done using the Kullback-Liebler measure, which is a restriction of Equation 1 to $g(\gamma) = 0$.

Related issues of tradeoffs in belief network construction are discussed in [5]. These dynamic network reformulation techniques can be used to identify the optimal resources devoted to network evaluation, and may help define the computation resource measures introduced in Section 5.1. These techniques may also be pertinent to facilitating the network construction approaches discussed here.

---

[9] Data for this AAP domain was briefly discussed in [24].

[10] This last criterion is similar to the Bayesian criterion presented in [7].

[11] Many latent parameters are rough subjective estimates. Further data collection and analysis is planned to rectify this problem.

# 8 CONCLUSIONS

This paper has proposed several approaches for constructing parsimonious TIDs for systems which evolve over time, where the state of the system during any time interval is modeled using a Bayesian network. Possible approaches to modeling the dynamic structure of the system have been examined, and the tradeoffs entailed in adopting particular approaches quantified using a variety of metrics. As an example, these techniques are applied to the medical management of acute abdominal pain.

In addition, this paper has proposed methods for selecting models with better predictive accuracy, and for trading off predictive accuracy for simpler models. Especially for complex domains such as temporal reasoning, limiting network size without compromising predictive accuracy too much can play an important role in ensuring computational tractability.

**Acknowledgements:** Dr. J.R. Clarke provided the medical expertise necessary for this research. Selecting a parsimonious subset of parameters for time series modeling was suggested to me by G. Rutledge. This work has also been influenced by discussions with D. Foster and M. Mintz, and by the anonymous referees.